\documentclass[preprint,12pt]{elsarticle}

\usepackage{amssymb}
\usepackage{amsmath}
\usepackage{subfigure}
\usepackage{amsfonts}
\usepackage{multirow}
\usepackage{booktabs}
\usepackage[switch]{lineno}
\usepackage{listings}
\usepackage{xcolor}

\usepackage{algorithm}
\usepackage{algpseudocode}

\usepackage{amsthm}  % 跟Definition有关的
\theoremstyle{definition}
\newtheorem{definition}{Definition}

\begin{document}

\begin{frontmatter}



\title{KnowPC: Knowledge-Driven Programmatic Reinforcement Learning for Zero-shot Coordination}

%% \affiliation[label2]

\author[a]{Yin Gu}
\author[a]{Qi Liu\corref{cor1}}
\cortext[cor1]{Corresponding author}
\ead{qiliuql@ustc.edu.cn}
\author[b]{Zhi Li}
\author[a]{Kai Zhang}

\affiliation[a]{organization={State Key Laboratory of Cognitive Intelligence, University of Science and Technology of China}, city={Hefei}, postcode={230000}, country={China}}
\affiliation[b]{organization={Shenzhen International Graduate School, Tsinghua University}, city={Shenzhen}, postcode={518000}, country={China}}

\begin{abstract}
	
Zero-shot coordination (ZSC) remains a major challenge in the cooperative AI field, which aims to learn an agent to cooperate with an unseen partner in training environments or even novel environments. In recent years, a popular ZSC solution paradigm has been deep reinforcement learning (DRL) combined with advanced self-play or population-based methods to enhance the neural policy's ability to handle unseen partners. Despite some success, these approaches usually rely on black-box neural networks as the policy function. However, neural networks typically lack interpretability and logic, making the learned policies difficult for partners (e.g., humans) to understand and limiting their generalization ability.
These shortcomings hinder the application of reinforcement learning methods in diverse cooperative scenarios.
In this paper, we suggest to represent the agent's policy with an interpretable program. 
Unlike neural networks, programs contain stable logic, but they are non-differentiable and difficult to optimize.
To automatically learn such programs, we introduce \textbf{Know}ledge-driven \textbf{P}rogrammatic reinforcement learning for zero-shot \textbf{C}oordination (KnowPC). We first define a foundational Domain-Specific Language (DSL), including program structures, conditional primitives, and action primitives. A significant challenge is the vast program search space, making it difficult to find high-performing programs efficiently. To address this, KnowPC integrates an \textbf{extractor} and an \textbf{reasoner}. The extractor discovers environmental transition knowledge from multi-agent interaction trajectories, while the reasoner deduces the preconditions of each action primitive based on the transition knowledge.
Together, they enable KnowPC to reason efficiently within an abstract space, reducing the trial-and-error cost. 
Finally, a program \textbf{synthesizer} generates desired programs based on the given DSL and the preconditions of the action primitives.
We choose the popular two-player cooperative game Overcooked as the experimental environment. Extensive experiments reveal the effectiveness of KnowPC, achieving comparable or superior performance compared to advanced DRL methods. Notably, when the environment layout changes, KnowPC continues to make stable decisions, whereas the DRL baselines fail.

\end{abstract}




\end{frontmatter}

%\linenumbers
%In the literature, many existing methods in group comparison~\cite{herbrich2007trueskill,huang2008ranking} focus on learning individual effects from outcomes of group comparisons. 

Collaboration between agents or between agents and humans is common in various scenarios, such as industrial robots~\cite{mukherjee2022survey,semeraro2023human,melo2019project}, game AI~\cite{bard2020hanabi, ashktorab2020human, barrett2017making,ribeiro2023teamster}, and autonomous driving~\cite{kyriakidis2019human,mariani2021coordination}. In these open scenarios, agents cannot anticipate the policies of the partners they will collaborate with, so they must be capable of cooperating with a wide range of unseen policies. This task is known as zero-shot coordination (ZSC)~\cite{hu2020other,li2023cooperative}.

In the literature, a prevailing approach to addressing ZSC has been deep reinforcement learning (DRL)~\cite{mnih2015human} coupled with improved self-play~\cite{tesauro1994td} or population-based methods~\cite{jaderberg2017population}.
Self-play~\cite{tesauro1994td,silver2017mastering} is a straightforward method where an agent plays with itself to iteratively optimize its policy. However, self-play tends to make the agent converge to specific behavior patterns, making it difficult to handle diverse partners~\cite{hu2020other}. 
Subsequently, various methods have been proposed to expose agents to diverse policies during training, aiming for better generalization.
Population-based training (BPT)~\cite{jaderberg2017population,carroll2019utility} maintains a diverse population and optimizes individual combinations iteratively.
Several improvements to BPT have been introduced~\cite{lupu2021trajectory,strouse2021collaborating,zhao2023maximum,lou2023pecan,li2023cooperative}. For instance, TrajeDi~\cite{lupu2021trajectory} encourages diversity in agents' trajectory distributions. MEP~\cite{zhao2023maximum} adds entropy of the average policy as an optimization objective to obtain a diverse population. 
%And COLE~\cite{li2023cooperative} identifies cooperative incompatibility and focuses on population subsets that are hard to cooperate with.
%
Recently, E3T~\cite{yan2023efficient} improved the self-play algorithm by mixing ego policy and random policy to promote diversity in partner policies.

Despite the success of existing work, there are still two major drawbacks. First, the neural network policies of DRL are not interpretable and are still considered a black box~\cite{rudin2019stop}. In cooperative decision-making scenarios, the interpretability of policies is important. Especially when cooperating with humans, if the agent's policies and behaviors can be understood by human partners, it can greatly increase human trust and promote cooperative efficiency~\cite{siu2021evaluation}. Secondly, neural policies lack inherent logic~\cite{cao2022galois} and mostly seek to fit the correlation between actions and expected returns, which makes their policies less robust and limits their generalization performance. 
This paper considers two forms of generalization tasks. One is to cooperate with a wide range of unseen partners~\cite{hu2020other,li2023cooperative,agapiou2022melting}, i.e., zero-shot coordination (ZSC), which is the task mainly considered in existing work. The other is to cooperate with unknown partners in unseen scenario layouts~\cite{cobbe2020leveraging,kirk2023survey}, which we name \textbf{ZSC+}. Layout variations are relatively common, such as different room layouts in different households or different layouts in games' maps. A good agent should be robust to such variations and be able to cooperate with unknown policies in any layout, rather than being limited to a specific layout. Clearly, ZSC+ is more challenging than ZSC and imposes higher requirements on the generalization performance of agents.

In stark contrast to neural policies, programmatically represented policies are fully interpretable~\cite{glanois2024survey} and possess stable logical rules, leading to better generalization performance. However, they are difficult to optimize or learn owing to their discrete and non-differentiable nature. To efficiently discover programs through trial and error, we propose \textbf{Know}ledge-Driven \textbf{P}rogrammatic reinforcement learning for zero-shot \textbf{C}oordination (KnowPC). In this paper, knowledge refers to the environment's transition rules, describing how elements in the environment change. KnowPC explicitly discovers and utilizes these transition rules to synthesize decision-logic-compliant programs as the agent's control policy. 
The training paradigm of KnowPC follows self-play, where in each episode, a programmatic policy is shared among all agents.
KnowPC integrates an \textbf{extractor}, an \textbf{reasoner}, and a program \textbf{synthesizer}. Specifically, the extractor identifies concise transition rules from multi-agent interaction trajectories and distinguishes between agent-caused transitions and spontaneous environmental transitions. The program synthesizer synthesizes programs based on a defined Domain-Specific Language (DSL). 
A significant technical challenge lies in the exponential increase of the program space with the program length. To tackle it, The reasoner uses the identified transition rules to determine the prerequisites of transitions, thereby establishing the preconditions for certain actions. This constrains the program search space of the synthesizer, improving search efficiency. The contributions of this paper are summarized as follows:

\begin{itemize}
	\item We introduce programmatic reinforcement learning in the ZSC task. Compared to neural policies, programmatic policies are fully interpretable and follow exact logical rules.
	
	\item The presented KnowPC explicitly extracts and leverages environmental knowledge and performs efficient reasoning in symbolic space to precisely synthesize programs that meet logical constraints.
	
	\item We consider a more complex task, ZSC+, which poses higher requirements on the generalization ability of agents. Extensive experiments on the well-established Overcooked~\cite{carroll2019utility} demonstrate that even with simple self-play training, KnowPC's policies outperform the existing methods in ZSC. Particularly, its generalization performance in ZSC+ far exceeds that of advanced baseline methods.
\end{itemize}

\section{Related Work}
\subsection{Zero-shot Coordination}

The mainstream approach to ZSC is to combine DRL and improved self-play or population-based training to develop policies that can effectively cooperate with unknown partners. Traditional self-play~\cite{tesauro1994td,silver2017mastering} methods control multiple agents by sharing policies and continuously optimizing the policy. However, self-play policies often perform poorly with unseen partners due to exhibiting a single behavior pattern. Other-play~\cite{hu2020other} exploits environmental symmetry to perturb agent policies and prevent them from degenerating into a single behavior pattern. Recent E3T~\cite{yan2023efficient} improved the self-play algorithm by mixing ego policy and random policy to promote diversity in partner policies and introduced an additional teammate modeling module to predict teammate action probabilities.
Population-based methods~\cite{jaderberg2017population,carroll2019utility,lupu2021trajectory,strouse2021collaborating,zhao2023maximum,lou2023pecan,li2023cooperative,charakorn2023generating} maintain a diverse population to train robust policies. Some advanced population-based methods enhance the diversity in different ways: FCP~\cite{strouse2021collaborating} preserves policies from different checkpoints during self-play training to increase population diversity, TrajeDi~\cite{lupu2021trajectory} maximizes differences in policy trajectory distributions, and MEP~\cite{zhao2023maximum} adds the entropy of the average policy in the population as an additional optimization objective. COLE~\cite{li2023cooperative} reformulates cooperative tasks as graphic-form games and iteratively learns a policy that approximates the best responses to the cooperative incompatibility distribution in the recent population.

Unlike previous work that focused on developing advanced self-play or population-based methods, we address this problem from the perspective of policy representation. By using programs with logical structure instead of black-box neural networks, we enhance the generalization ability of agents.

\subsection{Programmatic Reinforcement Learning}

Programmatic reinforcement learning represents its policies using interpretable symbolic languages~\cite{glanois2024survey}, including decision trees~\cite{ernst2005tree,bastani2018verifiable,silver2020few,silva2020optimization}, state machines~\cite{inala2020synthesizing}, and mathematical expressions~\cite{landajuela2021discovering,guo2023efficient}. The main challenge in programmatic reinforcement learning is the need to search for programs in a vast, non-differentiable space. Based on their learning or search methods, programmatic reinforcement learning approaches can be categorized into three types: imitation learning-based, differentiable architecture, and search-based.

Imitation learning-based methods~\cite{bastani2018verifiable,verma2018programmatically,verma2019imitation,inala2020synthesizing} first train a DRL policy to collect trajectory data, then use this data to learn interpretable programmatic policies.
Differentiable architecture methods~\cite{silva2020optimization,icct-rss-22,guo2023efficient,qiu2022programmatic} typically use an actor-critic framework~\cite{peters2008natural}, where the actor is a differentiable program or decision tree, and the critic is a neural network. Since raw programs are not differentiable, these methods use relaxation techniques to make the program structure differentiable. 
For instance, ICCTs~\cite{icct-rss-22} use sigmoid functions to compute the probabilities of visiting the left and right child nodes of a decision tree node, and recursively compute the probabilities of each leaf node in the tree, thus making the entire decision tree differentiable. Similarly, PRL~\cite{qiu2022programmatic} also uses sigmoid functions to compute the probabilities of left and right branches. However, the program structures in these methods are not very flexible, as they only allow if-else-then branches and do not permit sequential execution logic.

Because programs are discrete and difficult to optimize using gradient-based methods, a more straightforward approach is to search for the desired programs to use as policies. Search-based methods include genetic algorithms~\cite{koza1994genetic,katoch2021review,canaan2018evolving,carvalho2024reclaiming,moraes2023choosing}, Monte Carlo Tree Search (MCTS)~\cite{coulom2006efficient,kocsis2006bandit,medeiros2022can}, and DRL. DSP~\cite{landajuela2021discovering} uses DRL policies to output discrete mathematical expression tokens as control strategies, using risk-seeking gradient descent to optimize policy parameters. $\pi$-light~\cite{gu2024pi} predefines part of the program structure and then uses MCTS to search for the remaining parts of the program. A notable variant of search-based methods is LEAPS~\cite{trivedi2021learning}, which first learns a continuous program embedding space for discrete programs, then searches for continuous vectors in this space using the cross-entropy method~\cite{de2005tutorial}, and decodes them into discrete programs. 
Subsequent HPRL~\cite{liu2023hierarchical} improved its search method.

However, the aforementioned approaches do not extract and utilize environmental transition knowledge to accelerate the learning of programmatic policies. In contrast, KnowPC can infer the logical rules that programs must follow based on discovered transition knowledge.

\section{Preliminary}

\subsection{Environment}

\begin{figure} [htp]
	\centering
	\includegraphics[width=0.76\columnwidth]{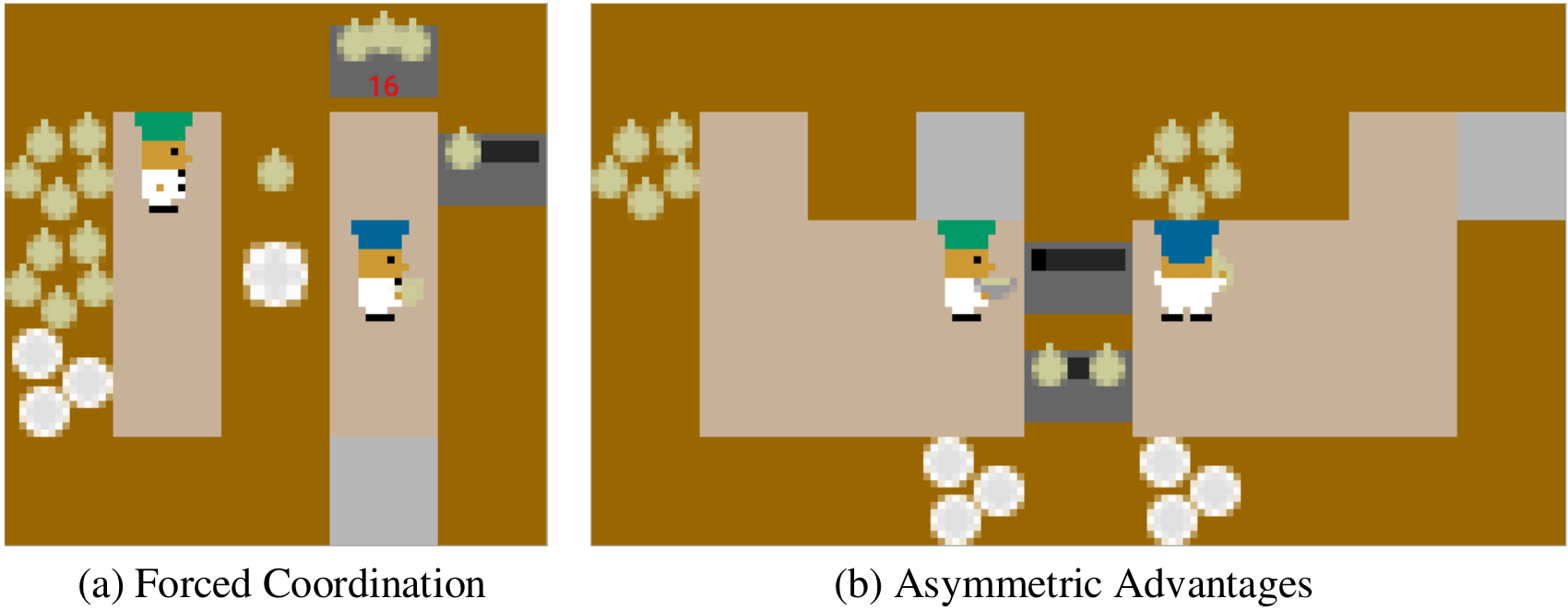}
	\caption{Illustration of Overcooked environment. We choose two layouts \textbf{Forced Coordination} and \textbf{Asymmetric Advantages} for demonstration.}
	\label{env}
\end{figure}

We choose the well-established multi-agent coordination suite Overcooked~\cite{carroll2019utility} as the experimental environment. Overcooked is a grid environment where agents independently control two chefs to make soup and deliver it. Figure \ref{env} shows two layouts of the environment. There are three types of objects in the environment: dishes, onions, and soup, along with four types of interaction points: onion dispenser, dish dispenser, counter, and pot. The chefs need to place three onions into a pot, wait for it to cook for 20 time steps, and then use an empty dish to serve the soup from the pot. After delivering the soup, all agents receive a reward of 20. Additionally, chefs can place any object they are holding onto an empty counter or pick up an object from a non-empty counter (provided the chef's hands are empty). Counters and pots are stateful interaction points, while onion dispensers and dish dispensers are stateless interaction points.

The two chefs on the field share the same discrete action space: up, down, left, right, noop, and ``interact". In each episode, an agent is randomly assigned to control one of the chefs. We refer to the controlled chef as the \emph{player} and the other chef as the \emph{teammate}. The teammate may be controlled by another unknown agent or a human. When trained in a self-play manner, the teammate is controlled by a copy of the current agent.

As introduced in early work~\cite{carroll2019utility,yan2023efficient}, in Overcooked, agents need to learn how to navigate, interact with objects, pick up the right objects, and place them in the correct locations. Most importantly, agents need to effectively cooperate with unseen agents.

\subsection{Cooperative Multi-agent MDP}

Two-player Cooperative Markov Decision Process: A two-player Markov Decision Process (MDP) is defined by a tuple $\langle \mathcal{S}, \mathcal{A}_1, \mathcal{A}_2, \mathcal{T}, \mathcal{R} \rangle$. Here, $\mathcal{S}$ is a finite state space, and $\mathcal{A}_1$ and $\mathcal{A}_2$ are the action spaces of the two agents, which we assume to be the same. The transition function $\mathcal{T}$ maps the current state and all agents' actions to the next state. The reward function $\mathcal{R}$ determines a real-valued reward based on the current state and all agents' actions, and this reward is shared among the agents. $\pi_1$ and $\pi_2$ are the policies of the two agents. Their goal is to maximize the cumulative reward: $\sum_{t=1}^H R(s^t, a^t_1, a^t_2)$, where $a^t_1 \sim \pi_1(s^t)$ and $a^t_2 \sim \pi_2(s^t)$. Here, $H$ is a finite time horizon.

\section{KnowPC Method}

\begin{figure} [htp]
	\centering
	\includegraphics[width=0.8\columnwidth]{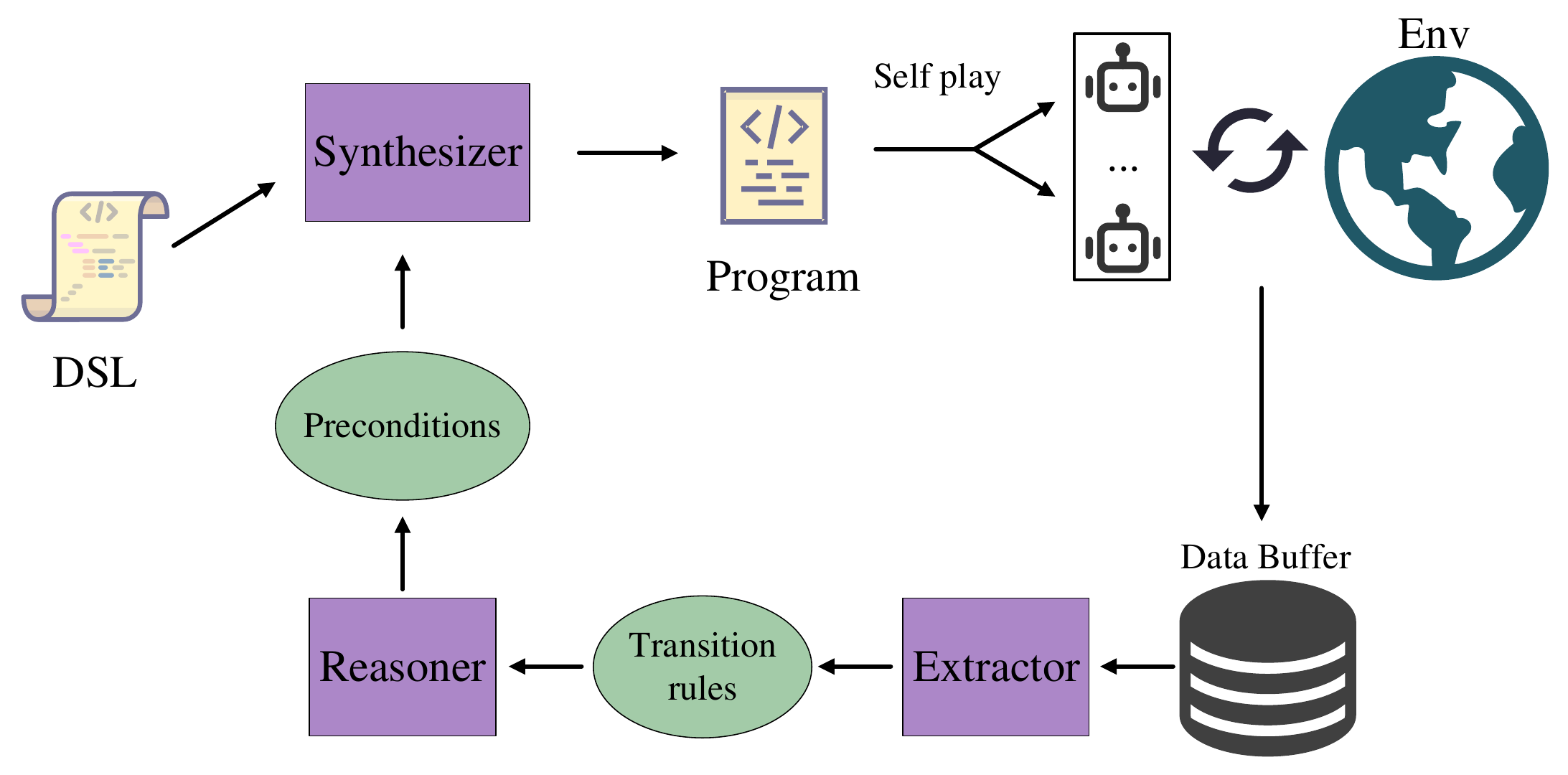}
	\caption{The overall framework of KnowPC. The data buffer is maintained and continues to increase in the learning loop.}
	\label{framework}
\end{figure}

In this section, we introduce the proposed KnowPC. As shown in Figure~\ref{framework}, KnowPC includes an extractor, a reasoner, and a program synthesizer. In KnowPC, the program serves as a decentralized policy to control one game character. During training, we adopt a simple self-play method where the programmatic policy is shared among all agents. Both agents explore with the $\epsilon$-greedy strategy~\cite{sutton2018reinforcement}, meaning they choose a random action with probability $\epsilon$ and choose the action output by the program with probability  $1-\epsilon$. The extractor aims to extract the environment's transition knowledge from multi-agent interaction trajectories. The reasoner uses these transition knowledge to infer the preconditions of action primitives to guide program synthesis. 

Several previous studies~\cite{lyu2019sdrl, jin2022creativity, zhuo2021creativity, liuintegrating} have used learned transition knowledge to improve the learning efficiency of DRL. This is often achieved by providing intrinsic rewards to DRL agents, which help address sparse-reward and hierarchical tasks. Instead of introducing additional intrinsic rewards for training agents, we infer the preconditions of action primitives to directly synthesize symbolic policies.

\subsection{Domain-Specific Language}

\begin{figure}
	\begin{minipage}{0.7\linewidth}
		\begin{align*}
			\ \ \ \ \ \ \ \text{Program}\ E &:= \left[\textbf{IT}_1, \textbf{IT}_2, ..., \right]. \\
			\textbf{IT} &:= \textbf{if}\ \mathcal B\ \textbf{then}\ A \\
			\mathcal B &:= B_1\ \text{and}\ B_2\ \text{and}\ ... \\
		\end{align*}
	\end{minipage}
	\begin{minipage}{1\linewidth}
		\caption{The domain-specific language for constructing our programs. \textbf{IT} is a module that contain \textbf{if}-\textbf{then} structure. $A$ and $B$ are action primitive and condition primitive. $\mathcal B$ is a conjunction of conditions. An element of $E$ is \textbf{IT}.}
		\label{DSL}
	\end{minipage}
\end{figure}
\begin{figure}
	%	\fbox{   % 加个框
		\begin{minipage}{1\linewidth}
			\begin{align*}
				\hat B &:= \texttt{HoldEmpty} \big |\ \texttt{HoldOnion} \big |\  \texttt{HoldDish} \big |\  \texttt{HoldSoup} \\ 
				& \quad \big |\ \texttt{ExServing} \big |\ \texttt{ExOnionDisp} \big |\ \texttt{ExDishDisp} \ \\
				& \quad \big |\ \texttt{ExOnionCounter} \big |\ \texttt{ExDishCounter} \big |\ \texttt{ExSoupCounter} \big |\ \texttt{ExEmptyCounter} \\
				& \quad \big |\ \texttt{ExIdlePot} \big |\ \texttt{ExReadyPot} \\
				B &:= \hat B\ |\ \text{not}\ \hat B \\
				% ====================== 分割线 ======================
				A &:= \texttt{GoIntServing}\ \big |\ \texttt{GoIntOnionDisp} \big |\ \texttt{GoIntDishDisp} \ \\
				& \quad \big|\ \texttt{GoIntOnionCounter} \big|\ \texttt{GoIntDishCounter} \\ 
				& \quad \big|\ \texttt{GoIntSoupCounter} \big|\ \texttt{GoIntEmptyCounter} \\
				& \quad \big|\ \texttt{GoIntIdlePot} \big |\ \texttt{GoIntReadyPot} \\
			\end{align*}
		\end{minipage}
		%	}
	\begin{minipage}{1\linewidth}
		\caption{Definitions of condition primitive $A$ and action primitive $B$. A vertical bar $|$ indicates choice.}
		\label{AB}
	\end{minipage}
\end{figure}

Our policies are entirely composed of interpretable programmatic language. In this subsection, we formally define the DSL that constructs our programs. As shown in Figure \ref{DSL} and Figure \ref{AB}, the DSL includes control flows (e.g., \textbf{IT} modules, sequential execution module $E$), condition primitives $B$, and action primitives $A$. In an $\textbf{IT}$ module, $\mathcal B$ can be referred to as the \emph{precondition} of $A$. The action primitive $A$ will only be executed when $\mathcal B$ is true.

The program structure allowed by the DSL is not fixed. For instance, any number of $\textbf{IT}$ modules can be added to $E$, or there can be any number of conditions $B$ in $\mathcal B$. For implementation simplicity, the program adopts a list-like structure~\cite{letham2015interpretable}. Each element in the list is an $\textbf{IT}$ module, and each $\textbf{IT}$ module can have multiple conditions. Although we have disabled the nesting of $\textbf{IT}$ modules, the expressiveness of such programs is sufficient. Multiple nested $\textbf{IT}$ can be equivalently replaced by adding conditions to $\mathcal B$.

Next, we provide detailed explanations for all condition primitives and action primitives shown in Figure \ref{AB}. Condition primitives will return a boolean value based on the state of the environment. 
They can be divided into two categories: player-related features and interaction-point-related features. \texttt{HoldEmpty}, \texttt{HoldOnion}, \texttt{HoldDish}, and \texttt{HoldSoup} are player-related features. For example, \texttt{HoldOnion} indicates whether the controlled player is holding an onion. Conditions starting with \texttt{Ex} indicate whether a certain type of interaction point exists on the field and can be reached by the controlled player. For instance, \texttt{ExOnionCounter} represents that there is a counter with an onion on it and that the player can reach it. In the DSL, \texttt{Serving} refers to the serving station, where players need to deliver the soup to earn a reward. \texttt{OnionDisp} and \texttt{DishDisp} refer to the onion dispenser and dish dispenser, respectively, while \texttt{OnionCounter} refers to a counter with onions on it. \texttt{IdlePot} indicates a pot waiting for onions, and \texttt{ReadyPot} indicates a pot with ready-to-serve soup. 
The DSL does not explicitly incorporate the teammate's state, as their behavior can be reflected through the state of other interaction points.

Action primitives are a type of high-level action that controls the player to move to the closest interaction point and interact with it. For example, \texttt{GoIntOnionDisp} means the player will move to the closest onion dispenser and interact with it (i.e., pick up an onion). \texttt{GoIntServing} indicates the player will move to the closest serving station and deliver the soup. Due to the gap between high-level actions and low-level environment actions, we introduce a low-level controller to transform high-level actions into low-level environment actions. Since the environment is represented as a grid and policy interpretability is a primary focus, we simply use the BFS pathfinding algorithm as the low-level controller. At each timestep, the low-level controller returns an environment action to control the player.
In continuous control environments, we can further train a goal-conditioned RL agent~\cite{nasiriany2019planning,liu2022goal} to achieve such low-level control.

% 【BFS inherently understands what the obstacles are, and teammates are also considered obstacles.

Please note that certain conditions must be met for interactions to occur. For example, if the player is holding something, interacting with a non-empty counter has no effect; if the player does not have soup, interacting with the serving station has no effect. When the prerequisites are met and the agent takes appropriate actions, the states of the agent and the interaction point will change.

\subsection{Extractor}

Extractor aims to mine environment transition rules from multi-agent interaction trajectories.
There are various types of interaction points in the environment, and each type may have several instances. Additionally, there are two roles: player and teammate. The agent's actions may change the state of the player or the interaction points, while the environment also undergoes spontaneous changes (e.g., the cooking time of food in the pot continuously increases). The goal of the extractor is to uncover concise transition rules that describe the complete dynamics of the environment. The main challenge in extracting transition rules is determining which transitions are caused by the agent itself (rather than by the teammate) and which transitions are spontaneous.

We focus on the player and interaction points in the environment. Suppose there are $N$ elements in the environment (elements include players and interaction points), and the information of element $i$ at time $t$ and $t+1$ can be represented as $I^t_i$ and $I^{t+1}_i$. For readability, we remove the $t$-related superscripts and use $I_i$ and $I^\prime_i$ instead. We use $K(I_i)$ to denote the type of element $i$, which can be either a player or an interaction point. If $i$ is an interaction point, there is an additional feature $KI(I_i)$ indicating its type, such as the counter or the pot. $S(I_i)$ represents the state of element $i$. For example, for a player, the state could be holding an onion, while for a pot, its state includes the number of onions and the cooking time. Interaction points also have an additional relative position marker $Pos(I_i)$, indicating the relative position of element $i$ to the player. `on' means the player is on the same tile as the interaction point, `face' means the player is facing the interaction point, and `away' refers to any other situation that is neither `on' nor `face'. Position markers are mutually exclusive.

By comparing the state changes of an element between two successive time steps, we can determine which element has changed. $C$ records such change of a element, where $C = (I, I^\prime),\ S(I) \ne S(I^\prime)$. For stateless interaction points $i$, we also record them regardless of whether they have changed, so $C = (I, I^\prime)$.
A single time step of environmental transition $T$ includes one or more $C$, $T = \{C_i\}$.

Given multi-agent interaction trajectories $[(s^1, a^1_1, a^1_2), (s^2, a^2_1, a^2_2), \ldots]$, we can derive the transition and action sequences $[(T^1, a^1), (T^2, a^2), \ldots]$. Here, $a$ represents the player's actions, and we do not consider the teammate's actions. The goal of the extractor is to identify the transitions caused by the player, denoted as $T_p$, and those caused by the environment spontaneously, denoted as $T_s$. $T_p$ and $T_s$ should be concise and not include irrelevant $C$.

At each time step, the agent observes three types of transitions: transitions caused by itself, transitions caused by teammates, and spontaneously occurring transitions. We will describe how to reveal these three types of transitions in the following sections.

\subsubsection{Player-caused Transitions}

Extractor determines whether a transition is caused by the player's actions based on the action statistics of the transition. Intuitively, if the action probability distribution is flat, it means that the transition will occur regardless of the actions taken by the agent. In other words, the transition is likely not caused by the agent's actions. Conversely, if the action probability distribution is concentrated, the transition is likely caused by the player.

Since our transitions are symbolic, we can directly count the number of actions of each transition to calculate the action probability $p$. The entropy of the action probability $p$ is defined as follows:

\begin{equation}
	H(p) = -\sum_{a \in \mathcal A}P(a)\log P(a)
\end{equation}

where $P(a)$ represents the action probability, and $H(p)$ is the entropy. The smaller the entropy, the more concentrated the probability distribution. If the $H(p)$ of a transition is relatively small, it is very likely caused by the player. Here, we introduce a threshold $\delta$ to filter out transitions with entropy greater than $\delta$. The found player-caused transitions are denoted by $T_p$, and the most frequent action $a$ of the $T_p$ is also recorded. We use $D_p$ to represent the set of $T_p$ and $a$, $D_p = \{(T_p, a)\}$.

The found $T_p$ may contain irrelevant $C$. For example, a $C$ caused by the teammate is observed by the player and cannot be distinguished based on $H(p)$. To remove redundant $T_p$ in $D_p$, we compare all $T_p$ pairwise and use shorter transitions to exclude longer transitions that contain it. For instance, if there are two $T_p$, $T_1 = \{C_1, C_2\}$ and $T_2 = \{C_1, C_2, C_3\}$, $T_2$ contains $T_1$ and has an extra $C_3$. According to $T_1$, we know that $C_1$ and $C_2$ are caused by the player, while $C_3$ in $T_2$ is not. Therefore, $T_2$ is redundant and should be excluded.

\subsubsection{Teammate-caused Transitions}

If the player and the teammate have the same functionalities in the environment, and they would follow the same transition rules. Given the $T_p$ set, we can shift to the teammate's perspective to identify teammate-caused transition within a transition. If their functionalities differ, we would need to separately identify player-caused and teammate-caused transitions.

\subsubsection{Spontaneously Occurring Transitions}
The transitions observed by the agent at any given moment include self-caused transitions, teammate-caused transitions, and spontaneously occurring transitions. The complete transition excluding the player-caused and teammate-caused transitions, leaves the environment's spontaneous transitions $T_s$. We use $D_s$ to represent the set of $T_s$, $D_s = \{T_s\}$.

\subsection{Reasoner}

The environment's transition rules describe how information about the player and interaction points changes. Based on this information, the reasoner can construct a transition graph, where the nodes include element information and actions. By traversing this graph, the reasoner should infer the preconditions for executing certain action primitives. Different from previous works~\cite{zhang2024proagent} that inform the LLM agent~\cite{chang2024survey} of preconditions of action primitives, our algorithm can automatically infer such preconditions.

\begin{figure} [htp]
	\centering
	\includegraphics[width=0.36\columnwidth]{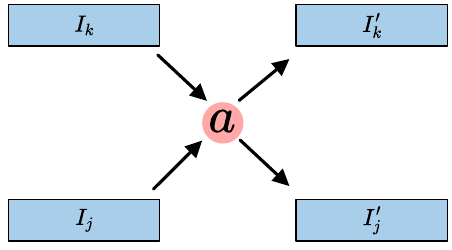}
	\caption{Illustration of a simple transition. }
	\label{simple_G}
\end{figure}

First, we introduce two concepts \emph{conjunctive conditions} and \emph{result} through a simple example. 
Suppose $j$ and $k$ are two elements, they have a related transition $T$ and a corresponding action $a$, where $T = \{C_j, C_k\} = \{(I_j,I_j^\prime),(I_k, I_k^\prime)\}$. We can describe it with a logical expression: $I_j \land I_k \overset{a}{\rightarrow} I_j^\prime \land I_k^\prime $. A corresponding transition graph can be drawn, as shown in Figure \ref{simple_G}. $I_j$ and $I_k$ both point to $a$, and $a$ points to $I_j^\prime$ and $I_k^\prime$. The meaning of this graph is that when both $I_j$ and $I_k$ are satisfied, executing action $a$ will result in $I_j^\prime$ and $I_k^\prime$. In the graph, $I_j$ and $I_k$ are the prerequisites of the transition, both of which must be met for the transition to occur, and $I_j^\prime$ and $I_k^\prime$ are the results of this $T$.

\begin{definition}
	Since there exists a $T$ and the prerequisites of $T$ are $I_j$ and $I_k$, $I_j$ and $I_k$ are each other's \emph{conjunctive conditions}.
\end{definition}

\begin{definition}
	The subsequent nodes of the action node are $I_j^\prime$ and $I_k^\prime$, so $I_j^\prime$ and $I_k^\prime$ are the \emph{results} of $I_j$. Similarly, $I_j^\prime$ and $I_k^\prime$ are also the \emph{results} of $I_k$.
\end{definition}

Next, we detail the reasoning algorithm. We first construct a transition graph $G$ using the extractor's outputs, $D_p$ and $D_s$. The logical expression of the transition in $D_p$ is similar to $I_j \land I_k \overset{a}{\rightarrow} I_j^\prime \land I_k^\prime$. For $T_s$ in $D_s$, we get $I_j \rightarrow I_j^\prime$ or $I_j \land I_k \rightarrow I_j^\prime \land I_k^\prime$, which means that these transitions can occur without any action. 
Note that we do not limit the number of prerequisites and results in a transition. 
The nodes in the graph are of three types: player nodes, which record the player's information; interaction point nodes, which record the interaction point's information; and action nodes, which record the environment's actions. 
In the transition graph, we do not distinguish between actions taken by the player or the teammate, as they have identical functionalities in the environment.

\begin{figure} [htp]
	\centering
	\includegraphics[width=1.0\columnwidth]{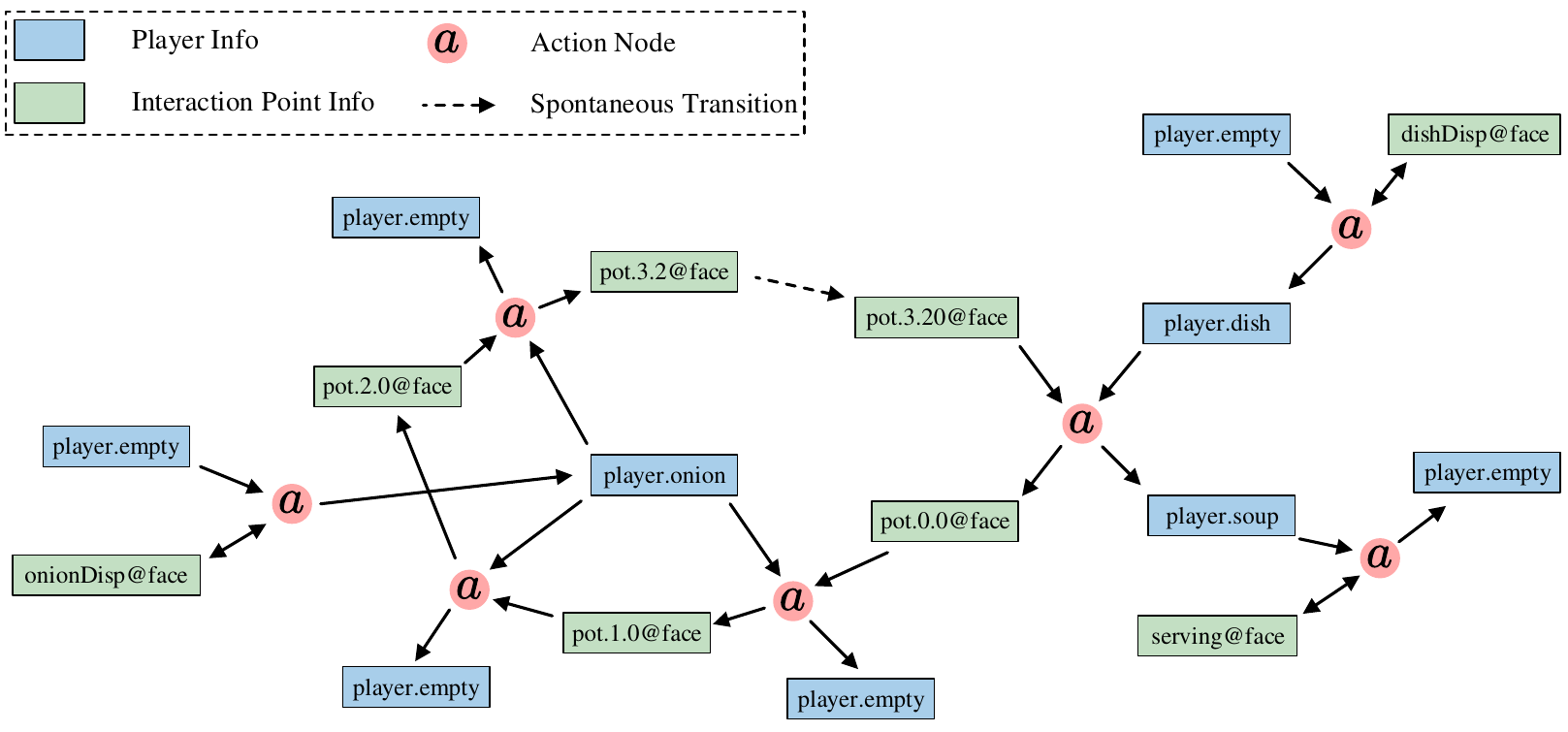}
	\caption{Transition graph constructed by the reasoner. For better representation, we use a string to fully describe an element's information. Here are some examples: 
	If $K(I)$ is `player' and $S(I)$ is `onion', the element is noted as `player.onion'.
	If $KI(I)$ is `dish dispenser' and $Pos(I)$ is `face', the element is noted as `dishDisp@face' (with the position identifier after `@').
	If $KI(I)$ is `pot', with 2 onions and 0 cooking time in the pot, and $Pos(I)$ is `face', the element is noted as `pot.2.0@face'. 
	A bidirectional arrow indicates that a node is both a prerequisite and a result of a transition.
	For clarity of presentation, nodes related to the table have been removed, and all instances of `player.empty' are considered different nodes.
	}
	\label{big_G}
\end{figure}

In Figure \ref{big_G}, we show a transition graph constructed by reasoner on the \textbf{Asymmetric Advantages} layout for illustration.
Assuming there exists a mapping that can convert element information into condition primitives and action primitives. 
e.g., `player.onion' can be mapped to \texttt{HoldOnion}. `serving@face' can be mapped to \texttt{ExServing} as a conditional primitive and to \texttt{GoIntServing} as an action primitive. We use $M_c(\cdot)$ to represent the mapping function from $I$ to conditional primitives, and $M_a(\cdot)$ to represent the mapping function from $I$ to action primitives.

\begin{algorithm}
	\caption{Reasoning Algorithm}
	\begin{algorithmic}[1]
		\State \textbf{Given:} Transition graph $G$
		\State \textbf{Output:} Preconditions for $M_a(I)$ of interaction point nodes $I$ in $G$
		\For{each interaction point node $I$ in $G$} 		\Comment{Single-step Reasoning}
		\For{$k$ in conjunctive conditions (CC) of $I$}
		\State Add $M_c(k)$ to the preconditions of $I$
		\EndFor
		\EndFor
		\For{each interaction point node $I$ in $G$}		\Comment{Multi-step Reasoning}
		\State \textit{Successors} = Results of $I$
		\For{each $j$ in \textit{Successors}}
		
		\For{$k$ in conjunctive conditions (CC) of $j$}
		\If{$k$ is not mutually exclusive with CC of $I$}
		\State Add $M_c(k)$ to the preconditions of $I$
		\EndIf
		\EndFor
		\EndFor
		\EndFor
	\end{algorithmic}
	\label{alg}
\end{algorithm}

During single-step reasoning, the reasoner focuses on interaction point nodes $I$ in the transition graph and identifies their conjunctive conditions as preconditions for $M_a(I)$. For example, for `onionDisp@face', its conjunctive condition is `player.empty'. This means the precondition for the player to go to the onion dispenser is that the player's hands are empty; otherwise, the interaction with the onion dispenser will not occur. Thus, the precondition for \texttt{GoIntOnionDisp} is \texttt{HoldEmpty}.

In multi-step reasoning, the reasoner considers the conjunctive conditions of the results of each interaction point node $I$. The preconditions for $M_a(I)$ include some of the conjunctive conditions of its results. For instance, the result of `onionDisp@face' is `player.onion', and the conjunctive condition for `player.onion' is \texttt{ExIdlePot}. The purpose of the player fetching an onion is to place it into an idle pot. If there is no idle pot available on the field, the player will either wait for an idle pot to appear or place the onion on the table. Therefore, the precondition for \texttt{GoIntOnionDisp} should include \texttt{ExIdlePot}.

The complete algorithm is described in Algorithm \ref{alg}. We detail both single-step reasoning and multi-step reasoning. Since the player's states are mutually exclusive, we exclude mutually exclusive conditions (line 10). Additionally, there is an induction step where condition primitives or action primitives with the same mapped name are aggregated together.

\subsection{Program synthesizer}

The program synthesizer will synthesize programs that conform to a given DSL. However, due to the vast program space, directly searching for high-performing programs within the given space is highly inefficient. To overcome this difficulty, our program synthesizer leverages the output of the reasoner, specifically the preconditions for each action primitive. These preconditions can be used to guide the synthesizer in generating reasonable programs, significantly reducing the search space.

As mentioned in the DSL subsection, our program structure adopts a list-like program. Each \textbf{IT} module in the program has variable $A$ and $\mathcal{B}$, which need to be determined by the synthesizer. We implement the synthesizer using a genetic algorithm~\cite{koza1994genetic,katoch2021review}, which includes selection and crossover operations. Initially, programs are randomly generated as the initial population. In each iteration, the crossover operation randomly selects parent programs and exchanges their program fragments to synthesize offspring. During the selection operation, programs with higher cumulative rewards in self-play are retained. 
We exclude the mutation step because it might alter the preconditions of action primitives, causing them to violate the requirements inferred by the reasoner. Additionally, given a state, it is possible that none of the $\mathcal B$ conditions in a program's \textbf{IT} modules are satisfied, resulting in an empty output. To prevent it, we append a random action at the end of every program.

After the genetic algorithm completes the search, we evaluate the discovered programs. During evaluation, we no longer use $\epsilon$-greedy exploration. 
To obtain a policy capable of cooperating with a variety of policies, we are inspired by population-based methods~\cite{carroll2019utility,li2023cooperative} to select a Pareto-optimal set from all the programs discovered.
The Pareto set is determined based on the training reward and the complexity of the programs, with complexity defined as the number of conditions $B$ in the program. Programs with higher cumulative rewards and lower complexity are considered better. This Pareto set includes diverse policies, such as those with the highest cumulative rewards and the simplest programs. A program is considered capable of handling diverse teammates if it cooperates well with each program in this set. Ultimately, the synthesizer outputs the program with the highest evaluation reward sum.

\subsubsection{Program Search Space Analysis}

The defined DSL contains 9 action primitives and 13 types of condition primitives (excluding the negation of conditions). Suppose a program uses 8 of these action primitives and 12 of these condition primitives, and the program has 8 \textbf{IT} modules, with each \textbf{IT} module containing up to 4 conditions. This results in approximately $1.64\times 10^{39}$ different possible programs.

The calculation process is as follows. First, we calculate the number of combinations of conditions in an \textbf{IT} module, selecting 1 to 4 conditions from 12 different ones, with each condition primitive being able to be negated:
$
C^1_{12}\times 2^1 + C^2_{12}\times 2^2 + C^3_{12}\times 2^3 + C^4_{12} \times 2^4= 9968.
$
Since there are 8 \textbf{IT} modules, each action in an \textbf{IT} module has 8 possible choices:
$(8\times9968)^ 8 \approx 1.64\times 10^{39}$. This calculation demonstrates that even for a moderately sized program, the potential combinatorial space is vast.

%\begin{equation}
%	\begin{split}
%	S_A & = \sum_{i\in T_A}w_i+\sum_{i\in T_A} \sum_{j\in T_A,i\ne j}\textbf h^T_{\text{1}}(\textbf v_i\odot \textbf v_j)\\
%	    & + \sum_{i\in T_A} \sum_{j\in T_B}\textbf h^T_{\text{2}} (\textbf p_i \odot \textbf c_j),
%	\end{split}
%\end{equation}
%where vector $\textbf{h}_1 \in \mathbb{R}^k$, $\textbf{h}_2 \in \mathbb{R}^k$ denotes neuron weights of the output laryer.

\section{Experiments}
\subsection{Experimental Setup} 

In this section, we evaluate KnowPC's ZSC and ZSC+ capabilities across multiple layouts in the Overcooked environment~\cite{carroll2019utility}. We compare the performance of KnowPC with six baselines: Self-Play (SP)~\cite{tesauro1994td,carroll2019utility}, Population-Based Training (PBT)~\cite{jaderberg2017population,carroll2019utility}, Fictitious Co-Play (FCP)~\cite{strouse2021collaborating}, Maximum-Entropy Population-Based Training (MEP)~\cite{zhao2023maximum}, Cooperative Open-ended Learning (COLE)~\cite{li2023cooperative}, and Efficient End-to-End Training (E3T)~\cite{yan2023efficient}. All of them use PPO~\cite{schulman2017proximal} as the RL algorithm and belong to DRL methods. Among them, SP and E3T are based on self-play, while the other algorithms are population-based, requiring the maintenance of a diverse population.

The parameter settings for KnowPC are consistent across different layouts. During training, the exploration probability $\epsilon$ is set to 0.3, and the threshold $\delta$ is set to 0.1. The genetic algorithm undergoes 50 iterations, with an initial population size of 200 and a subsequent population size maintained at 10.

It is worth noting that previous works have utilized shaped reward functions~\cite{carroll2019utility} to train agents, such as giving extra rewards for events like placing an onion into the pot, picking up a dish, or making a soup. This approach helps to accelerate convergence and improve performance.
In contrast, KnowPC is not sensitive to the reward function. We directly use a sparse reward function (i.e., only get a reward when delivering soup).
Additionally, the input encoding for DRL methods uses a lossless state encoding~\cite{carroll2019utility}, which includes multiple matrices with sizes corresponding to the environment grid size. In terms of state encoding, DRL has more complete information compared to KnowPC.

%\footnote{\scriptsize https://academictorrents.com/details/5c5deeb6cfe1c944044367d2e7465fd8bd2f4acf}

\subsection{ZSC Experiment Results}

\begin{figure*} [htp]
	\centering
	\includegraphics[width=0.72\columnwidth]{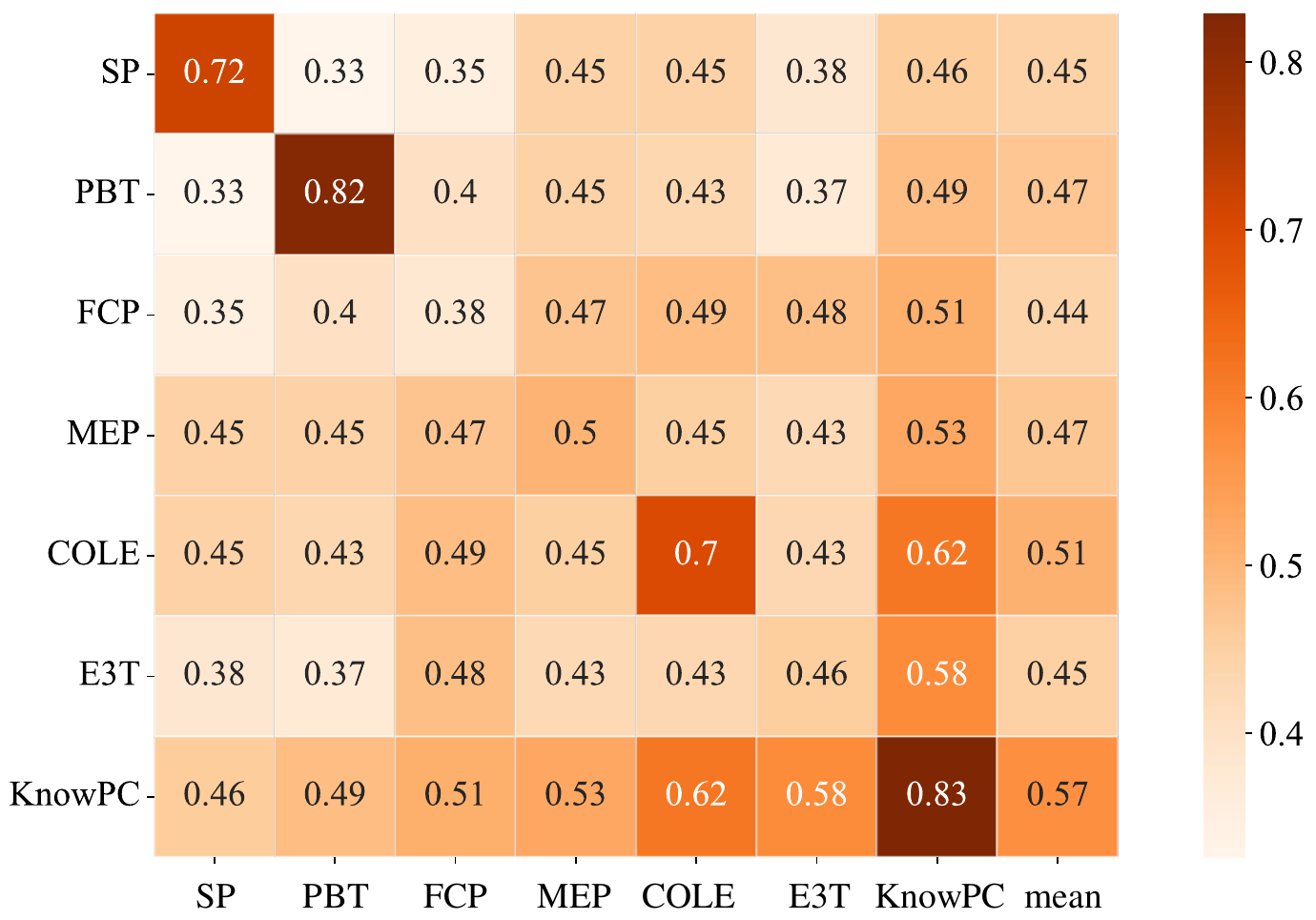}
	\caption{Normalized coordination rewards on the five layouts. The left 7x7 submatrix in this matrix is symmetric. The last column \textbf{mean} is the average of the first 7 columns.}
	\label{7v7}
\end{figure*}

\begin{figure*} [htp]
	\centering
	\includegraphics[width=0.78\columnwidth]{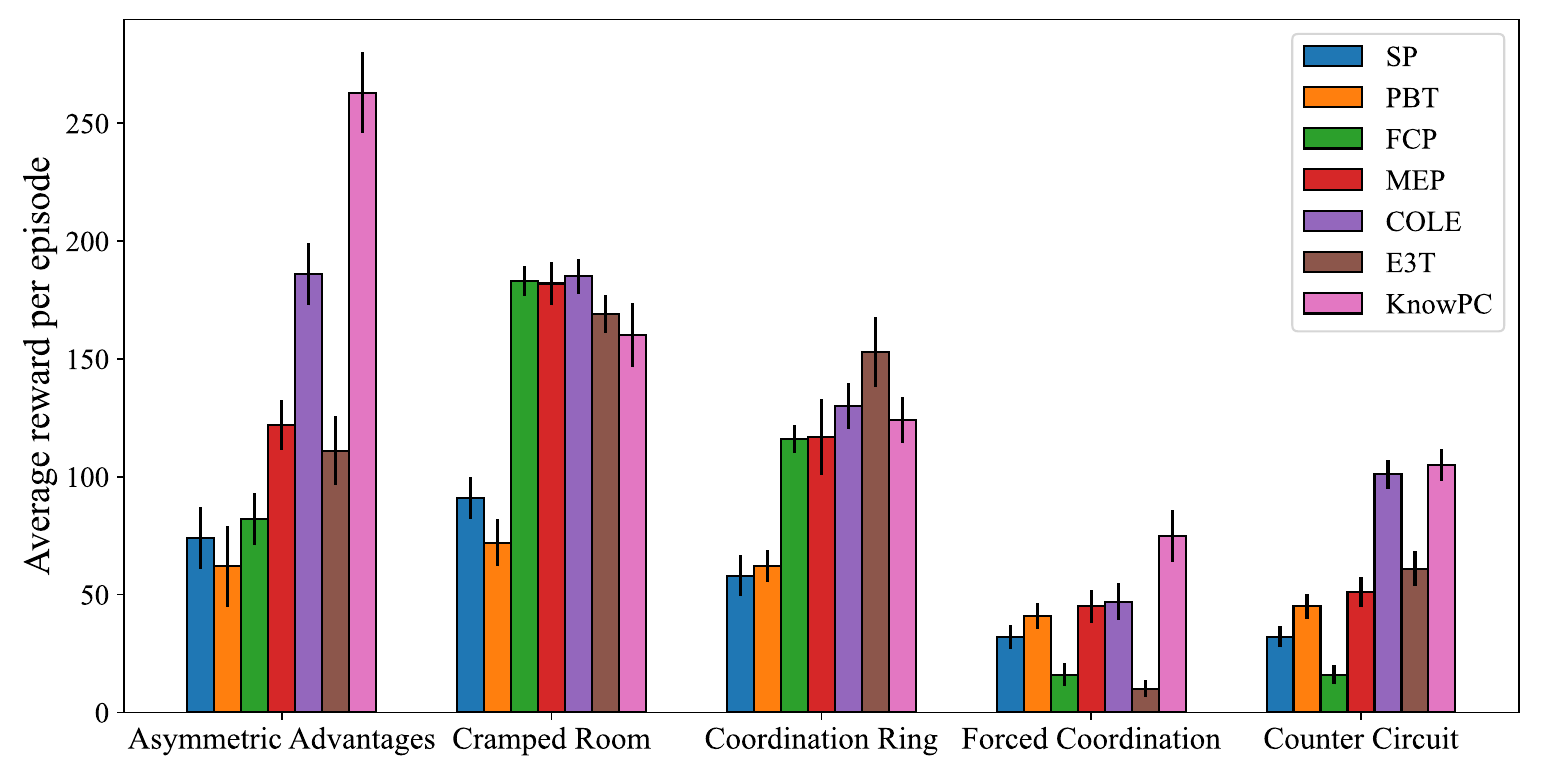}
	\caption{Average cumulative rewards when cooperating with behavior-cloned (BC) human proxies over 10 episodes. We show the mean rewards and standard error. Each reward bar represents the average reward of the agent and the BC model taking turns controlling both chefs.}
	\label{BC}
\end{figure*}

\subsubsection{Collaboration between RL Agents.}

We evaluate the ZSC capabilities of each method by letting each method's policies cooperate with each other.
During training, none of them can access the policies of other methods.
Figure \ref{7v7} shows the normalized cumulative reward values for pairwise cooperation of each method, averaged over 5 layouts. First, we observe that the self-play rewards of each method are generally higher than their rewards when cooperating with other unseen agents. Second, KnowPC's ZSC performance is higher than that of other baselines. Third, excluding self-play, each method achieves higher cumulative rewards when paired with KnowPC than others. This indicates KnowPC is the best companion except for themselves.

\subsubsection{Collaboration with Human Proxies}

Apart from collaborating with AI partners, an RL agent also needs to work with human partners. Following previous work~\cite{li2023cooperative,zhang2024proagent}, we use behavior-cloned models trained on human data as human proxies. Figure \ref{BC} presents the results on 5 layouts. KnowPC generally performs well overall. As noted in previous work, no method consistently outperforms the others. KnowPC achieves significantly higher results than the baselines in two layouts (Asymmetric Advantages and Forced Coordination). In two other layouts (Cramped Room and Coordination Ring), our results are slightly lower than the best baseline. Possibly because Cramped Room and Coordination Ring require more consideration and modeling of the teammates. Integrating agent modeling techniques~\cite{yan2023efficient} with KnowPC is a potential direction for future research.

\begin{table}
	\centering
	\caption{Comparison of the training time for each method. We report the training time on a single layout. The training times for the baselines are taken from their original papers.}
	\begin{tabular}{@{}cccccccc@{}}
		\toprule
		Method     & SP  & PBT & FCP& MEP &  COLE  & E3T & 	KnowPC  \\ \midrule
		Training Time(h) & 0.5  & 2.7   & 7.6 & 17.9  & 36  & 1.9  & \textbf{0.1}   \\ \bottomrule
	\end{tabular}
	\label{time}
\end{table}

\subsubsection{Training Efficiency Analysis}
Table \ref{time} shows the training times for all algorithms on a single layout. It can be observed that population-based methods (e.g., MEP, COLE) generally require more training time than self-play methods (e.g., SP, E3T). Our method is efficient. Benefiting from reasoning in the abstract space and not requiring extensive parameter optimization like DRL, it has the shortest training time among all methods. For instance, its training time is one 360th of previously advanced COLE and one 19th of the state-of-the-art E3T.

\subsection{Policy Interpretability}

\begin{lstlisting}[language=Python]
if ExIdlePot and HoldOnion: # First IT module
	GoIntIdlePot
if ExDishDisp and ExReadyPot and HoldEmpty: # Second 
	GoIntDishDisp
if ExReadyPot and HoldDish:
	GoIntReadyPot
if ExOnionDisp and ExIdlePot and not ExReadyPot and HoldEmpty: # Fourth
	GoIntOnionDisp
if ExEmptyCounter and not HoldEmpty and HoldDish:
	GoIntEmptyCounter
if ExReadyPot and ExDishCounter and HoldEmpty: # Sixth
	GoIntDishCounter
if ExIdlePot and ExOnionCounter and HoldEmpty:
	GoIntOnionCounter
if ExServing and HoldSoup: # Eighth
	GoIntServing
RandomAct
\end{lstlisting}

Listing 1 illustrates a program found by KnowPC on the Counter Circuit environment. Unlike the DRL policy, the program is fully interpretable, with transparent decision logic. For example, the logic expressed by the fourth \textbf{IT} module is that if there is an onion dispenser and an idle pot, and no ready pot, the player will go to the onion dispenser. This is reasonable because if there is no idle pot in the scene, it is futile for the player to go to the onion dispenser to get an onion (since only idle pots need onions). Handling the ready pot is prioritized over the idle pot, because handling the ready pot can get rewards more quickly, hence the $\mathcal B$ includes \texttt{not ExReadyPot}. In the sixth \textbf{IT} module, if there is a ready pot and a dish counter in the scene and the player’s hands are empty, the player will go to the dish counter. This is also a reasonable decision because dishes can only be used to serve soup from a ready pot, and obtaining a dish requires empty hands.

\subsection{ZSC+ Experiment Results}

% =========================== 三个layout  ===========================
\begin{figure*} [htp]
	\centering
	\includegraphics[width=1\columnwidth]{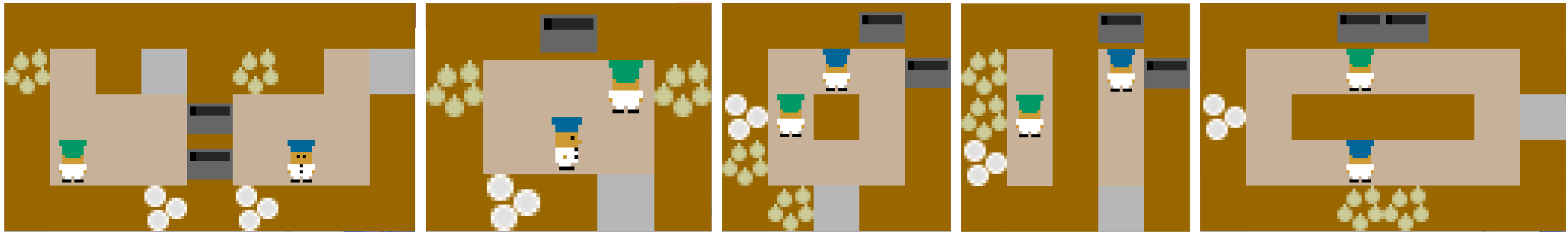}
	\caption{Original layouts in Overcooked. From left to right are \textbf{Asymmetric Advantages}, \textbf{Cramped Room}, \textbf{Coordination Ring}, \textbf{Forced Coordination}, and \textbf{Counter Circuit}.}
	\label{maps}
\end{figure*}
\begin{figure*} [htp]  % 一组新的layout。与原始layout不一样的交互点用红框标出。
	\centering
	\includegraphics[width=1\columnwidth]{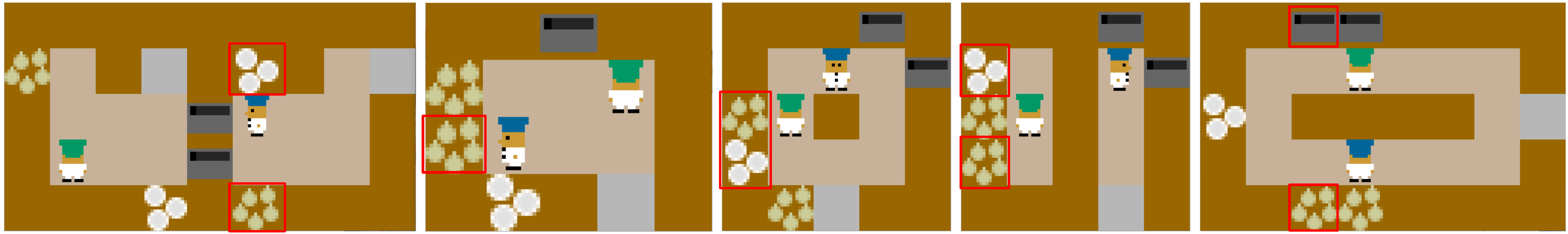}
	\caption{A new set of layouts. The Interaction points that are different from the original ones are marked with red boxes.}
	\label{maps_x}
\end{figure*}
\begin{figure*} [htp]
	\centering
	\includegraphics[width=1\columnwidth]{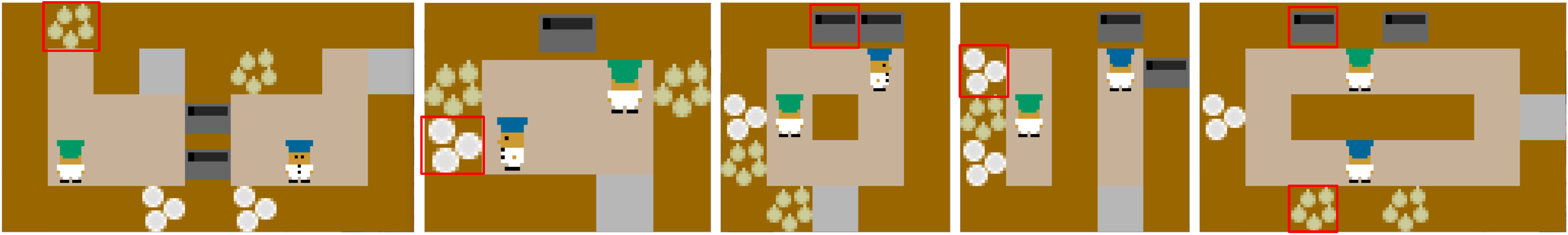}
	\caption{Another set of new layouts. The interaction points that are different from the original ones are marked with red boxes.}
	\label{maps_y}
\end{figure*}
% =========================== 三个layout  ===========================

\begin{figure*} [htp]
	\centering
	\includegraphics[width=0.72\columnwidth]{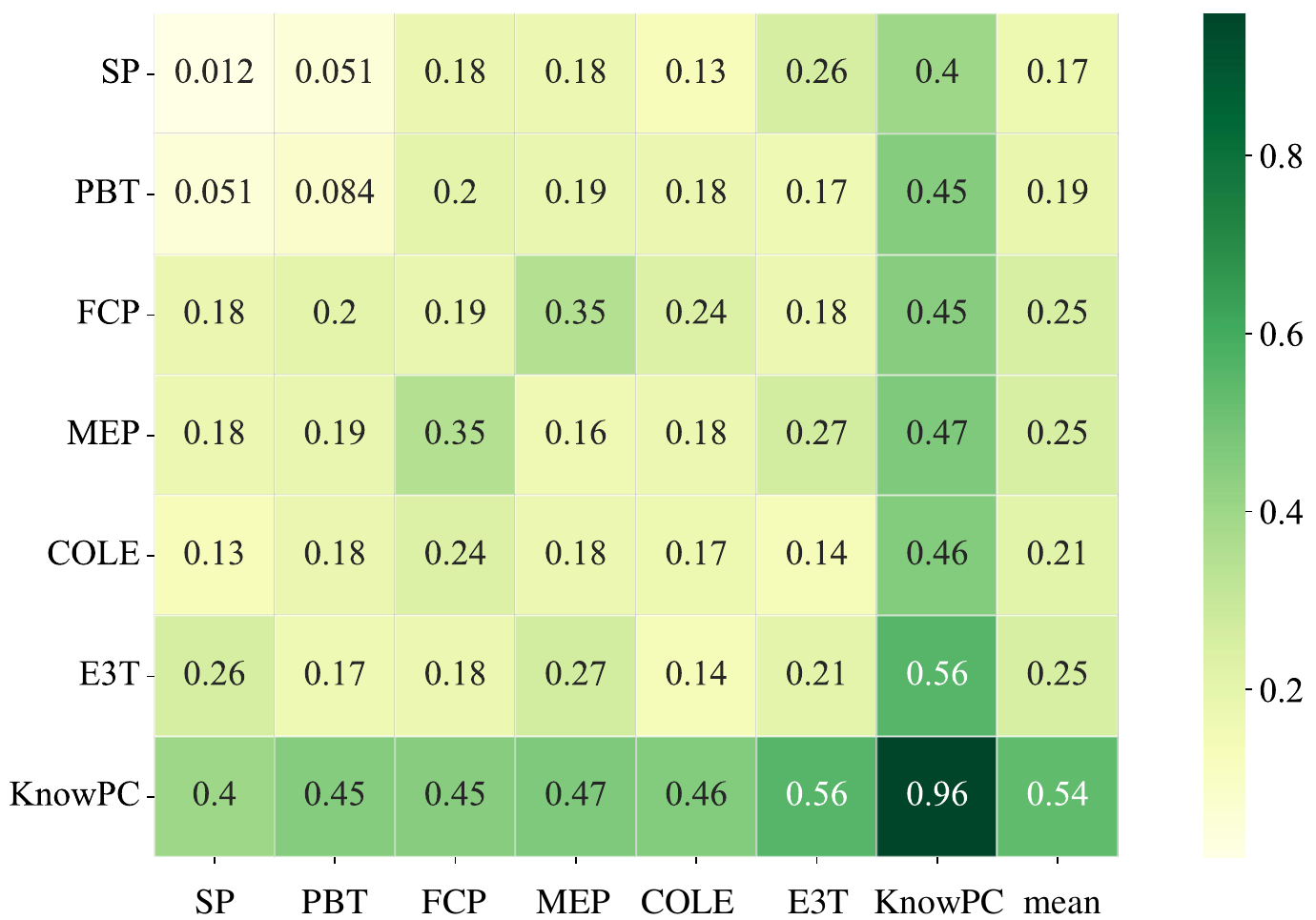}
	\caption{Normalized coordination rewards on the five new layouts.}
	\label{7v7_x}
\end{figure*}

\begin{figure*} [htp]
	\centering
	\includegraphics[width=0.72\columnwidth]{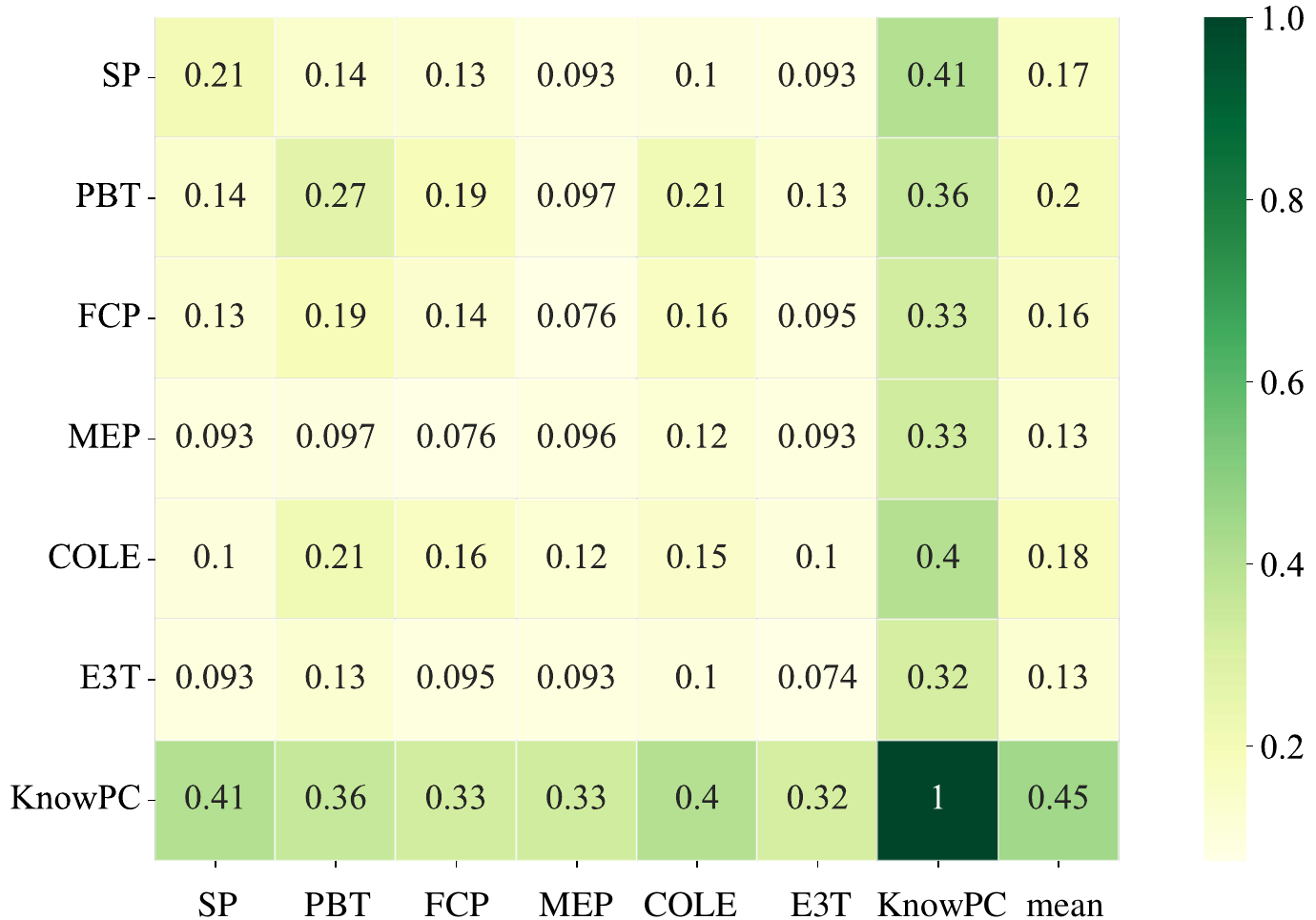}
	\caption{Normalized coordination rewards on the five new layouts.}
	\label{7v7_y}
\end{figure*}

Layout variations are common, as rooms in different buildings often have different layouts. A good RL policy should be robust to these layout changes. We are particularly interested in the ZSC+ capability of various methods, so we made some adjustments to the original Overcooked layout. As shown in Figures \ref{7v7_x} and \ref{7v7_y}, we created two new layout groups. Compared to the original layout (Figure \ref{maps}), the positions of several interaction points are slightly different. These Layout changes do not alter the dynamics of the environment but adjust the positions of some interaction points. This can test whether the agent has learned the underlying logic of decision-making rather than memorizing a fixed position. 

To verify the generalization ability of each method, we directly apply the policies trained on the original layout to the new layout without additional training. Figures \ref{7v7_x} and \ref{7v7_y} show the cooperation reward scores of each method in the two new layouts. The results show that the performance of all baselines declined to some extent. However, the performance of KnowPC was significantly better than the other methods, with its corresponding reward column exceeding those of the other baselines. This demonstrates that KnowPC's programmatic policies are robust.

We hypothesize that the failure of DRL is due to its policies being represented by an over-parameterized neural network. Neural networks struggle to deduce accurate decision logic purely from rewards, making them unable to handle unseen situations. In contrast, concise programs act as a form of policy regularization, making them more robust to unseen agents and layouts.

\subsection{Ablation Study}

\begin{table}
	\centering
	\setlength{\tabcolsep}{12pt}
	\caption{Rewards for each variant of KnowPC. The cumulative rewards are averaged over 5 different runs, with the variance shown in parentheses.}
	\begin{tabular}{@{}cccc@{}}
		\toprule
		Layout                & KnowPC  & KnowPC-M & PC   \\ \midrule
		Asymmetric Advantages & 445.6 (4.963)  & 346.0 (89.16)   & 0 (0)  \\ 
		Cramped Room          & 194.8 (6.145)  & 144.0 (108.22)  & 0 (0)  \\
		Coordination Ring     & 152.8 (37.62)  & 132.0 (39.81)   & 0 (0)  \\ 
		Forced Coordination   & 223.6 (111.87) & 61.2 (109.86)   & 0 (0)  \\ 
		Counter Circuit       & 158.4 (11.76)  & 116.8 (55.82)   & 0 (0)  \\
		\bottomrule
	\end{tabular}
	\label{abl}
\end{table}

To validate the effectiveness of the knowledge-driven extractor and reasoner, we first visualized a transition graph. As shown in Figure~\ref{big_G}, the extractor identified the correct environment transition rules. The transition graph clearly shows how the states of players and interaction points change and which changes are caused by actions and which are not.

To further verify the effectiveness of the reasoner, we conducted some ablation experiments. First, we removed the reasoner from KnowPC, resulting in a method called PC. Without the guidance of preconditions, PC's synthesizer searches for programs solely through a genetic algorithm. Next, to test the necessity of multi-step reasoning in the reasoner, we removed the multi-step reasoning step, keeping only single-step reasoning. This new method is called KnowPC-M.

As shown in Table~\ref{abl}, we report the self-play rewards for each variant's policy. By comparing PC and KnowPC, we found that the reasoner and environmental knowledge are indispensable to the entire framework. Without any of them, we cannot infer the preconditions of the action primitive, which leads to the complete failure of PC. KnowPC-M performed worse than KnowPC across all layouts, demonstrating that multi-step reasoning allows the reasoner to consider longer-term implications and provide more comprehensive preconditions for each action primitive.

\section{Conclusion and Future Work}

In this paper, we propose Knowledge-driven Programmatic reinforcement learning for zero-shot Coordination (KnowPC), which deploys programs as agent control policies. KnowPC extracts and generalizes knowledge about the environment and performs efficient reasoning in the symbolic space to synthesize programs that meet logical constraints. KnowPC integrates an extractor to uncover transition knowledge of the environment, and a reasoner to deduce the preconditions of action primitives based on this transition knowledge. The program synthesizer then searches for high-performing programs based on the given DSL and the deduced preconditions. 
Compared to DRL-based methods, KnowPC stands out for its interpretability, generalization performance, and robustness to sparse rewards. Its policies are fully interpretable, making it easier for humans to understand and debug them.
Empirical results on Overcooked demonstrate that, even in sparse reward settings, KnowPC can achieve superior performance compared to advanced baselines. Moreover, when the environment changes, program-based policies remain robust, while DRL baselines experience significant performance declines.

The limitation of our method is that it requires the definition of some basic conditional primitives and action primitives. This abstraction of the task necessitates a small amount of expert knowledge. Large language models might be adept at defining these conditional and action primitives, which we leave for future work.

%\section{Acknowledgements}
%This research was partially supported by grants from the National Key Research and Development Program of China (No. 2018YFC0832101), the National Natural Science Foundation of China (Grants No. 61922073 and 61672483), and the Foundation of State Key Laboratory of Cognitive Intelligence (Grant No. iED2020-M004).


\bibliographystyle{elsarticle-num}
\bibliography{ref.bib}

\begin{thebibliography}{10}
\expandafter\ifx\csname url\endcsname\relax
  \def\url#1{\texttt{#1}}\fi
\expandafter\ifx\csname urlprefix\endcsname\relax\def\urlprefix{URL }\fi
\expandafter\ifx\csname href\endcsname\relax
  \def\href#1#2{#2} \def\path#1{#1}\fi

\bibitem{mukherjee2022survey}
D.~Mukherjee, K.~Gupta, L.~H. Chang, H.~Najjaran, A survey of robot learning
  strategies for human-robot collaboration in industrial settings, Robotics and
  Computer-Integrated Manufacturing 73 (2022) 102231.

\bibitem{semeraro2023human}
F.~Semeraro, A.~Griffiths, A.~Cangelosi, Human--robot collaboration and machine
  learning: A systematic review of recent research, Robotics and
  Computer-Integrated Manufacturing 79 (2023) 102432.

\bibitem{melo2019project}
F.~S. Melo, A.~Sardinha, D.~Belo, M.~Couto, M.~Faria, A.~Farias, H.~Gamboa,
  C.~Jesus, M.~Kinarullathil, P.~Lima, et~al., Project inside: towards
  autonomous semi-unstructured human--robot social interaction in autism
  therapy, Artificial intelligence in medicine 96 (2019) 198--216.

\bibitem{bard2020hanabi}
N.~Bard, J.~N. Foerster, S.~Chandar, N.~Burch, M.~Lanctot, H.~F. Song,
  E.~Parisotto, V.~Dumoulin, S.~Moitra, E.~Hughes, et~al., The hanabi
  challenge: A new frontier for ai research, Artificial Intelligence 280 (2020)
  103216.

\bibitem{ashktorab2020human}
Z.~Ashktorab, Q.~V. Liao, C.~Dugan, J.~Johnson, Q.~Pan, W.~Zhang, S.~Kumaravel,
  M.~Campbell, Human-ai collaboration in a cooperative game setting: Measuring
  social perception and outcomes, Proceedings of the ACM on Human-Computer
  Interaction 4~(CSCW2) (2020) 1--20.

\bibitem{barrett2017making}
S.~Barrett, A.~Rosenfeld, S.~Kraus, P.~Stone, Making friends on the fly:
  Cooperating with new teammates, Artificial Intelligence 242 (2017) 132--171.

\bibitem{ribeiro2023teamster}
J.~G. Ribeiro, G.~Rodrigues, A.~Sardinha, F.~S. Melo, Teamster: Model-based
  reinforcement learning for ad hoc teamwork, Artificial Intelligence 324
  (2023) 104013.

\bibitem{kyriakidis2019human}
M.~Kyriakidis, J.~C. de~Winter, N.~Stanton, T.~Bellet, B.~van Arem,
  K.~Brookhuis, M.~H. Martens, K.~Bengler, J.~Andersson, N.~Merat, et~al., A
  human factors perspective on automated driving, Theoretical issues in
  ergonomics science 20~(3) (2019) 223--249.

\bibitem{mariani2021coordination}
S.~Mariani, G.~Cabri, F.~Zambonelli, Coordination of autonomous vehicles:
  Taxonomy and survey, ACM Computing Surveys (CSUR) 54~(1) (2021) 1--33.

\bibitem{hu2020other}
H.~Hu, A.~Lerer, A.~Peysakhovich, J.~Foerster, other-play for zero-shot
  coordination, in: International Conference on Machine Learning, PMLR, 2020,
  pp. 4399--4410.

\bibitem{li2023cooperative}
Y.~Li, S.~Zhang, J.~Sun, Y.~Du, Y.~Wen, X.~Wang, W.~Pan, Cooperative open-ended
  learning framework for zero-shot coordination, in: International Conference
  on Machine Learning, PMLR, 2023, pp. 20470--20484.

\bibitem{mnih2015human}
V.~Mnih, K.~Kavukcuoglu, D.~Silver, A.~A. Rusu, J.~Veness, M.~G. Bellemare,
  A.~Graves, M.~Riedmiller, A.~K. Fidjeland, G.~Ostrovski, et~al., Human-level
  control through deep reinforcement learning, nature 518~(7540) (2015)
  529--533.

\bibitem{tesauro1994td}
G.~Tesauro, Td-gammon, a self-teaching backgammon program, achieves
  master-level play, Neural computation 6~(2) (1994) 215--219.

\bibitem{jaderberg2017population}
M.~Jaderberg, V.~Dalibard, S.~Osindero, W.~M. Czarnecki, J.~Donahue, A.~Razavi,
  O.~Vinyals, T.~Green, I.~Dunning, K.~Simonyan, et~al., Population based
  training of neural networks, arXiv preprint arXiv:1711.09846 (2017).

\bibitem{silver2017mastering}
D.~Silver, J.~Schrittwieser, K.~Simonyan, I.~Antonoglou, A.~Huang, A.~Guez,
  T.~Hubert, L.~Baker, M.~Lai, A.~Bolton, et~al., Mastering the game of go
  without human knowledge, nature 550~(7676) (2017) 354--359.

\bibitem{carroll2019utility}
M.~Carroll, R.~Shah, M.~K. Ho, T.~Griffiths, S.~Seshia, P.~Abbeel, A.~Dragan,
  On the utility of learning about humans for human-ai coordination, Advances
  in neural information processing systems 32 (2019).

\bibitem{lupu2021trajectory}
A.~Lupu, B.~Cui, H.~Hu, J.~Foerster, Trajectory diversity for zero-shot
  coordination, in: International conference on machine learning, PMLR, 2021,
  pp. 7204--7213.

\bibitem{strouse2021collaborating}
D.~Strouse, K.~McKee, M.~Botvinick, E.~Hughes, R.~Everett, Collaborating with
  humans without human data, Advances in Neural Information Processing Systems
  34 (2021) 14502--14515.

\bibitem{zhao2023maximum}
R.~Zhao, J.~Song, Y.~Yuan, H.~Hu, Y.~Gao, Y.~Wu, Z.~Sun, W.~Yang, Maximum
  entropy population-based training for zero-shot human-ai coordination, in:
  Proceedings of the AAAI Conference on Artificial Intelligence, Vol.~37, 2023,
  pp. 6145--6153.

\bibitem{lou2023pecan}
X.~Lou, J.~Guo, J.~Zhang, J.~Wang, K.~Huang, Y.~Du, Pecan: Leveraging policy
  ensemble for context-aware zero-shot human-ai coordination, in: Proceedings
  of the International Joint Conference on Autonomous Agents and Multiagent
  Systems, AAMAS, Vol. 2023, International Foundation for Autonomous Agents and
  Multiagent Systems, 2023, pp. 679--688.

\bibitem{yan2023efficient}
X.~Yan, J.~Guo, X.~Lou, J.~Wang, H.~Zhang, Y.~Du, An efficient end-to-end
  training approach for zero-shot human-ai coordination, in: Proceedings of the
  37th International Conference on Neural Information Processing Systems, 2023,
  pp. 2636--2658.

\bibitem{rudin2019stop}
C.~Rudin, Stop explaining black box machine learning models for high stakes
  decisions and use interpretable models instead, Nature machine intelligence
  1~(5) (2019) 206--215.

\bibitem{siu2021evaluation}
H.~C. Siu, J.~Pe{\~n}a, E.~Chen, Y.~Zhou, V.~Lopez, K.~Palko, K.~Chang,
  R.~Allen, Evaluation of human-ai teams for learned and rule-based agents in
  hanabi, Advances in Neural Information Processing Systems 34 (2021)
  16183--16195.

\bibitem{cao2022galois}
Y.~Cao, Z.~Li, T.~Yang, H.~Zhang, Y.~Zheng, Y.~Li, J.~Hao, Y.~Liu, Galois:
  boosting deep reinforcement learning via generalizable logic synthesis,
  Advances in Neural Information Processing Systems 35 (2022) 19930--19943.

\bibitem{agapiou2022melting}
J.~P. Agapiou, A.~S. Vezhnevets, E.~A. Du{\'e}{\~n}ez-Guzm{\'a}n, J.~Matyas,
  Y.~Mao, P.~Sunehag, R.~K{\"o}ster, U.~Madhushani, K.~Kopparapu, R.~Comanescu,
  et~al., Melting pot 2.0, arXiv preprint arXiv:2211.13746 (2022).

\bibitem{cobbe2020leveraging}
K.~Cobbe, C.~Hesse, J.~Hilton, J.~Schulman, Leveraging procedural generation to
  benchmark reinforcement learning, in: International conference on machine
  learning, PMLR, 2020, pp. 2048--2056.

\bibitem{kirk2023survey}
R.~Kirk, A.~Zhang, E.~Grefenstette, T.~Rockt{\"a}schel, A survey of zero-shot
  generalisation in deep reinforcement learning, Journal of Artificial
  Intelligence Research 76 (2023) 201--264.

\bibitem{glanois2024survey}
C.~Glanois, P.~Weng, M.~Zimmer, D.~Li, T.~Yang, J.~Hao, W.~Liu, A survey on
  interpretable reinforcement learning, Machine Learning (2024) 1--44.

\bibitem{charakorn2023generating}
R.~Charakorn, P.~Manoonpong, N.~Dilokthanakul, Generating diverse cooperative
  agents by learning incompatible policies, in: The Eleventh International
  Conference on Learning Representations, 2023.

\bibitem{ernst2005tree}
D.~Ernst, P.~Geurts, L.~Wehenkel, Tree-based batch mode reinforcement learning,
  Journal of Machine Learning Research 6 (2005).

\bibitem{bastani2018verifiable}
O.~Bastani, Y.~Pu, A.~Solar-Lezama, Verifiable reinforcement learning via
  policy extraction, Advances in neural information processing systems 31
  (2018).

\bibitem{silver2020few}
T.~Silver, K.~R. Allen, A.~K. Lew, L.~P. Kaelbling, J.~Tenenbaum, Few-shot
  bayesian imitation learning with logical program policies, in: Proceedings of
  the AAAI Conference on Artificial Intelligence, Vol.~34, 2020, pp.
  10251--10258.

\bibitem{silva2020optimization}
A.~Silva, M.~Gombolay, T.~Killian, I.~Jimenez, S.-H. Son, Optimization methods
  for interpretable differentiable decision trees applied to reinforcement
  learning, in: International conference on artificial intelligence and
  statistics, PMLR, 2020, pp. 1855--1865.

\bibitem{inala2020synthesizing}
J.~P. Inala, O.~Bastani, Z.~Tavares, A.~Solar-Lezama, Synthesizing programmatic
  policies that inductively generalize, in: 8th International Conference on
  Learning Representations, 2020.

\bibitem{landajuela2021discovering}
M.~Landajuela, B.~K. Petersen, S.~Kim, C.~P. Santiago, R.~Glatt, N.~Mundhenk,
  J.~F. Pettit, D.~Faissol, Discovering symbolic policies with deep
  reinforcement learning, in: International Conference on Machine Learning,
  PMLR, 2021, pp. 5979--5989.

\bibitem{guo2023efficient}
J.~Guo, R.~Zhang, S.~Peng, Q.~Yi, X.~Hu, R.~Chen, Z.~Du, X.~Zhang, L.~Li,
  Q.~Guo, et~al., Efficient symbolic policy learning with differentiable
  symbolic expression, in: Thirty-seventh Conference on Neural Information
  Processing Systems, 2023.

\bibitem{verma2018programmatically}
A.~Verma, V.~Murali, R.~Singh, P.~Kohli, S.~Chaudhuri, Programmatically
  interpretable reinforcement learning, in: International Conference on Machine
  Learning, PMLR, 2018, pp. 5045--5054.

\bibitem{verma2019imitation}
A.~Verma, H.~Le, Y.~Yue, S.~Chaudhuri, Imitation-projected programmatic
  reinforcement learning, Advances in Neural Information Processing Systems 32
  (2019).

\bibitem{icct-rss-22}
R.~Paleja, Y.~Niu, A.~Silva, C.~Ritchie, S.~Choi, M.~Gombolay, Learning
  interpretable, high-performing policies for autonomous driving, in: Robotics:
  Science and Systems (RSS), 2022.

\bibitem{qiu2022programmatic}
W.~Qiu, H.~Zhu, Programmatic reinforcement learning without oracles, in: The
  Tenth International Conference on Learning Representations, 2022.

\bibitem{peters2008natural}
J.~Peters, S.~Schaal, Natural actor-critic, Neurocomputing 71~(7-9) (2008)
  1180--1190.

\bibitem{koza1994genetic}
J.~R. Koza, Genetic programming as a means for programming computers by natural
  selection, Statistics and computing 4 (1994) 87--112.

\bibitem{katoch2021review}
S.~Katoch, S.~S. Chauhan, V.~Kumar, A review on genetic algorithm: past,
  present, and future, Multimedia tools and applications 80 (2021) 8091--8126.

\bibitem{canaan2018evolving}
R.~Canaan, H.~Shen, R.~Torrado, J.~Togelius, A.~Nealen, S.~Menzel, Evolving
  agents for the hanabi 2018 cig competition, in: 2018 IEEE Conference on
  Computational Intelligence and Games (CIG), IEEE, 2018, pp. 1--8.

\bibitem{carvalho2024reclaiming}
T.~H. Carvalho, K.~Tjhia, L.~Lelis, Reclaiming the source of programmatic
  policies: Programmatic versus latent spaces, in: The Twelfth International
  Conference on Learning Representations, 2024.

\bibitem{moraes2023choosing}
R.~O. Moraes, D.~S. Aleixo, L.~N. Ferreira, L.~H. Lelis, Choosing well your
  opponents: how to guide the synthesis of programmatic strategies, in:
  Proceedings of the Thirty-Second International Joint Conference on Artificial
  Intelligence, 2023, pp. 4847--4854.

\bibitem{coulom2006efficient}
R.~Coulom, Efficient selectivity and backup operators in monte-carlo tree
  search, in: International conference on computers and games, Springer, 2006,
  pp. 72--83.

\bibitem{kocsis2006bandit}
L.~Kocsis, C.~Szepesv{\'a}ri, Bandit based monte-carlo planning, in: European
  conference on machine learning, Springer, 2006, pp. 282--293.

\bibitem{medeiros2022can}
L.~C. Medeiros, D.~S. Aleixo, L.~H. Lelis, What can we learn even from the
  weakest? learning sketches for programmatic strategies, in: Proceedings of
  the AAAI Conference on Artificial Intelligence, Vol.~36, 2022, pp.
  7761--7769.

\bibitem{gu2024pi}
Y.~Gu, K.~Zhang, Q.~Liu, W.~Gao, L.~Li, J.~Zhou, $\pi$-light: Programmatic
  interpretable reinforcement learning for resource-limited traffic signal
  control, in: Proceedings of the AAAI Conference on Artificial Intelligence,
  Vol.~38, 2024, pp. 21107--21115.

\bibitem{trivedi2021learning}
D.~Trivedi, J.~Zhang, S.-H. Sun, J.~J. Lim, Learning to synthesize programs as
  interpretable and generalizable policies, Advances in neural information
  processing systems 34 (2021) 25146--25163.

\bibitem{de2005tutorial}
P.-T. De~Boer, D.~P. Kroese, S.~Mannor, R.~Y. Rubinstein, A tutorial on the
  cross-entropy method, Annals of operations research 134 (2005) 19--67.

\bibitem{liu2023hierarchical}
G.-T. Liu, E.-P. Hu, P.-J. Cheng, H.-Y. Lee, S.-H. Sun, Hierarchical
  programmatic reinforcement learning via learning to compose programs, in:
  International Conference on Machine Learning, PMLR, 2023, pp. 21672--21697.

\bibitem{sutton2018reinforcement}
R.~S. Sutton, A.~G. Barto, Reinforcement learning: An introduction, MIT press,
  2018.

\bibitem{lyu2019sdrl}
D.~Lyu, F.~Yang, B.~Liu, S.~Gustafson, Sdrl: interpretable and data-efficient
  deep reinforcement learning leveraging symbolic planning, in: Proceedings of
  the AAAI Conference on Artificial Intelligence, Vol.~33, 2019, pp.
  2970--2977.

\bibitem{jin2022creativity}
M.~Jin, Z.~Ma, K.~Jin, H.~H. Zhuo, C.~Chen, C.~Yu, Creativity of ai: Automatic
  symbolic option discovery for facilitating deep reinforcement learning, in:
  Proceedings of the AAAI Conference on Artificial Intelligence, Vol.~36, 2022,
  pp. 7042--7050.

\bibitem{zhuo2021creativity}
H.~H. Zhuo, S.~Deng, M.~Jin, Z.~Ma, K.~Jin, C.~Chen, C.~Yu, Creativity of ai:
  Hierarchical planning model learning for facilitating deep reinforcement
  learning, arXiv preprint arXiv:2112.09836 (2021).

\bibitem{liuintegrating}
J.-C. Liu, C.-H. Chang, S.-H. Sun, T.-L. Yu, Integrating planning and deep
  reinforcement learning via automatic induction of task substructures, in: The
  Twelfth International Conference on Learning Representations.

\bibitem{letham2015interpretable}
B.~Letham, C.~Rudin, T.~H. McCormick, D.~Madigan, Interpretable classifiers
  using rules and bayesian analysis: Building a better stroke prediction model
  (2015).

\bibitem{nasiriany2019planning}
S.~Nasiriany, V.~Pong, S.~Lin, S.~Levine, Planning with goal-conditioned
  policies, Advances in neural information processing systems 32 (2019).

\bibitem{liu2022goal}
M.~Liu, M.~Zhu, W.~Zhang, Goal-conditioned reinforcement learning: Problems and
  solutions, arXiv preprint arXiv:2201.08299 (2022).

\bibitem{zhang2024proagent}
C.~Zhang, K.~Yang, S.~Hu, Z.~Wang, G.~Li, Y.~Sun, C.~Zhang, Z.~Zhang, A.~Liu,
  S.-C. Zhu, et~al., Proagent: building proactive cooperative agents with large
  language models, in: Proceedings of the AAAI Conference on Artificial
  Intelligence, Vol.~38, 2024, pp. 17591--17599.

\bibitem{chang2024survey}
Y.~Chang, X.~Wang, J.~Wang, Y.~Wu, L.~Yang, K.~Zhu, H.~Chen, X.~Yi, C.~Wang,
  Y.~Wang, et~al., A survey on evaluation of large language models, ACM
  Transactions on Intelligent Systems and Technology 15~(3) (2024) 1--45.

\bibitem{schulman2017proximal}
J.~Schulman, F.~Wolski, P.~Dhariwal, A.~Radford, O.~Klimov, Proximal policy
  optimization algorithms, arXiv preprint arXiv:1707.06347 (2017).

\end{thebibliography}

\end{document}